\documentclass[10pt,twocolumn,letterpaper]{article}

\usepackage{cvpr}
\usepackage{times}
\usepackage{epsfig}
\usepackage{graphicx}
\usepackage{amsmath}
\usepackage{amssymb}

\usepackage{color}
\definecolor{citecolor}{RGB}{119,185,0} 
\usepackage[pagebackref=true,breaklinks=true,letterpaper=true,colorlinks,citecolor=citecolor,bookmarks=false]{hyperref}

\usepackage{multirow}
\usepackage{cite}
\usepackage{floatrow}
\usepackage{caption}
\usepackage{comment}
\usepackage[dvipsnames,svgnames,x11names]{xcolor}
\newcommand{\tabincell}[2]{\begin{tabular}{@{}#1@{}}#2\end{tabular}}
\usepackage{pifont}

\cvprfinalcopy 

\def\cvprPaperID{16} 

\newlength\savewidth\newcommand\shline{\noalign{\global\savewidth\arrayrulewidth
  \global\arrayrulewidth 1pt}\hline\noalign{\global\arrayrulewidth\savewidth}}

\ifcvprfinal\fi
\begin{document}

\title{Joint Discriminative and Generative Learning for Person Re-identification}

\author{
  Zhedong Zheng${^{1,2}}$\thanks{Work done during an internship at NVIDIA Research.} \quad Xiaodong Yang${^1}$ \quad Zhiding Yu${^1}$\\
  Liang Zheng${^3}$ \quad Yi Yang${^2}$ \quad Jan Kautz${^1}$  \\ 
  $^1$NVIDIA ~$^2$CAI, University of Technology Sydney ~$^3$Australian National University\\
}

\maketitle

\begin{abstract}
Person re-identification (re-id) remains challenging due to significant intra-class variations across different cameras. Recently, there has been a growing interest in using generative models to augment training data and enhance the invariance to input changes. The generative pipelines in existing methods, however, stay relatively separate from the discriminative re-id learning stages. Accordingly, re-id models are often trained in a straightforward manner on the generated data. In this paper, we seek to improve learned re-id embeddings by better leveraging the generated data. To this end, we propose a joint learning framework that couples re-id learning and data generation end-to-end. Our model involves a generative module that separately encodes each person into an appearance code and a structure code, and a discriminative module that shares the appearance encoder with the generative module. By switching the appearance or structure codes, the generative module is able to generate high-quality cross-id composed images, which are online fed back to the appearance encoder and used to improve the discriminative module. The proposed joint learning framework renders significant improvement over the baseline without using generated data, leading to the state-of-the-art performance on several benchmark datasets.
\end{abstract}

\begin{figure}[t]
\begin{center}
   \includegraphics[width=0.915\linewidth]{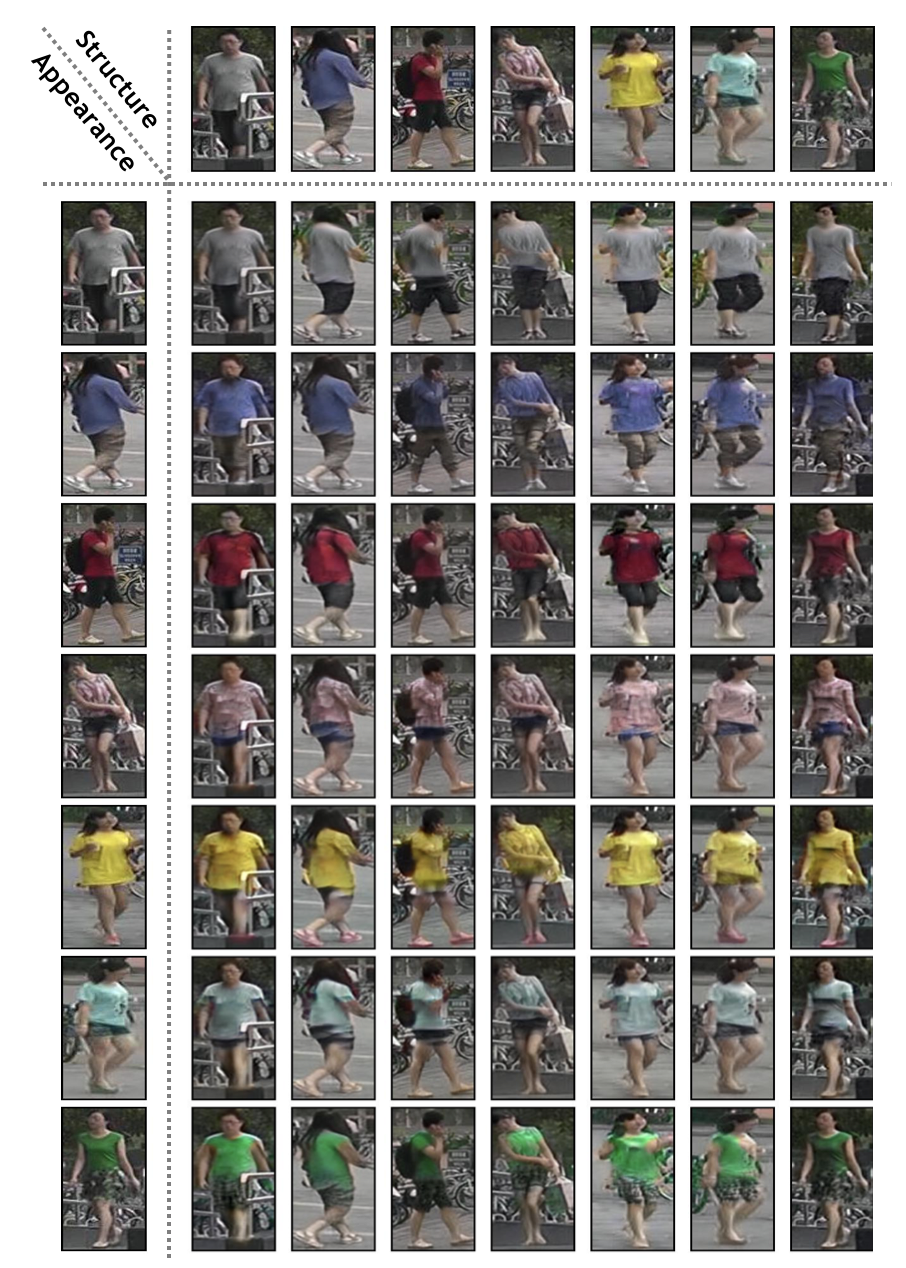}
\end{center}\vspace{-.2in}
\caption{Examples of generated images on Market-1501 by switching appearance or structure codes. Each row and column corresponds to different appearance and structure.}
\label{fig:rainbow}
\end{figure}

\section{Introduction}
Person re-identification (re-id) aims to establish identity correspondences across different cameras. It is often approached as a metric learning problem \cite{zheng2016survey}, where one seeks to retrieve images containing the person of interest from non-overlapping cameras given a query image. This is challenging in the sense that images captured by different cameras often contain significant intra-class variations caused by the changes in background, viewpoint, human pose, etc.
As a result, designing or learning representations that are robust against intra-class variations as much as possible has been one of the major targets in person re-id. 

Convolutional neural networks (CNNs) have recently become increasingly predominant choices in person re-id thanks to their strong representation power and the ability to learn invariant deep embeddings.
Current state-of-the-art re-id methods widely formulate the tasks as deep metric learning problems \cite{zheng2016discriminatively, hermans2017defense}, or use classification losses as the proxy targets to learn deep embeddings \cite{zheng2016survey, li2017person, sun2017beyond, wu2019progressive, zheng2018pedestrian, tang@cityflow}. 
To further reduce the influence from intra-class variations, a number of existing methods 
adopt part-based matching or ensemble to explicitly align and compensate the variations \cite{su2017pose, zhao2017spindle, wei2017glad, suh2018part, zheng2018pedestrian}. 

\setlength{\tabcolsep}{9pt}
\begin{table}
\caption{Description of the information encoded in the latent appearance and structure spaces.}\vspace{-.1in}{
\begin{tabular}{c|c}
\shline
Appearance Space & Structure Space \\
\hline 
 \tabincell{c}{clothing/shoes color,\\
 texture and style,\\
 other id-related cues, etc.}
& \tabincell{c}{body size, hair, carrying,\\ 
pose, background,\\ 
position, viewpoint, etc.}\\
\shline
\end{tabular}}
\label{table:info}
\end{table}

Another possibility to enhance robustness against input variations is to let the re-id model potentially ``see'' these variations (particularly intra-class variations) during training. With recent progress in the generative adversarial networks (GANs) \cite{goodfellow2014generative}, generative models have become appealing choices to introduce additional augmented data for free \cite{zheng2017unlabeled}. Despite the different forms, the general considerations behind these methods are ``realism'': generated images should possess good qualities to close the domain gap between synthesized scenarios and real ones; and ``diversity'': generated images should contain sufficient diversity to adequately cover unseen variations. 
Within this context, some prior works have explored unconditional GANs and human pose conditioned GANs~\cite{zheng2017unlabeled, huang2018multi, qian2017pose, ge2018fdgan, liu2018pose} to generate pedestrian images to improve re-id learning. 
However, a common issue behind these methods is that their generative pipelines are typically presented as standalone models, which are relatively separate from the discriminative re-id models. Therefore, 
the optimization target of a generative module may not be well aligned with the re-id task, limiting the gain from generated data.

In light of the above observation, we propose a learning framework that jointly couples discriminative and generative learning in a unified network called \textbf{DG-Net}. Our strategy towards achieving this goal is to introduce a generative module, of which encoders decompose each pedestrian image into two latent spaces: an \textbf{appearance} space that mostly encodes appearance and other identity related semantics; and a \textbf{structure} space that encloses geometry and position related structural information as well as other additional variations. 
We refer to the encoded features in the space as ``codes''. The properties captured by the two latent spaces are summarized in Table~\ref{table:info}. The appearance space encoder is also shared with the discriminative module, serving as a re-id learning backbone. This design leads to a single unified 
framework that subsumes these interactions between generative and discriminative modules: (1) the generative module produces synthesized images that are taken to refine the appearance encoder online; (2) the encoder, in turn, influences the generative module with improved appearance encoding; and (3) both modules are jointly optimized, given the shared appearance encoder.

We formulate the image generation as switching the appearance or structure codes between two images. Given any pairwise images with the same/different identities, one is able to generate realistic and diverse intra/cross-id composed images by manipulating the codes. An example of such composed image generation on Market-1501~\cite{zheng2015scalable} is shown in Figure~\ref{fig:rainbow}. Our design of the generative pipeline not only leads to high-fidelity generation, but also yields substantial diversity given the 
combinatorial compositions of existing identities. Unlike the unconditional GANs~\cite{zheng2017unlabeled, huang2018multi}, our method allows more controllable generation with better quality. Unlike the pose-guided generations~\cite{qian2017pose,ge2018fdgan, liu2018pose}, our method does not require any additional auxiliary data, but takes the advantage of existing intra-dataset pose variations as well as other diversities beyond pose. 

This generative module design specifically serves for our discriminative module to better make use of the generated data. For one pedestrian image, by keeping its appearance code and combining with different structure codes, we can generate multiple images that remain clothing and shoes but change pose, viewpoint, background, etc. As demonstrated in each row of Figure~\ref{fig:rainbow}, these images correspond to the same clothing dressed on different people. 
To better capture such composed cross-id information, we introduce the ``primary feature learning'' via a dynamic soft labeling strategy. Alternatively, we can keep one structure code and combine with different appearance codes to produce various images, which maintain the pose, background and some identity related fine details but alter clothes and shoes. As shown in each column of Figure~\ref{fig:rainbow}, these images form an interesting simulation of the same person wearing different clothes and shoes. This creates an opportunity for further mining the subtle identity attributes that are independent of clothing, such as carrying, hair, body size, etc. Thus, we propose the complementary ``fine-grained feature mining'' to learn additional subtle identity properties. 

To our knowledge, this work provides the first framework that is able to end-to-end integrate discriminative and generative learning in a single unified network for person re-id. Extensive qualitative and quantitative experiments show that our image generation compares favorably against the existing ones, and more importantly, our re-id accuracy consistently outperforms the competing algorithms by large margins on several benchmarks. 


\begin{figure*}[t]
\begin{center}
   \includegraphics[width=0.97\linewidth]{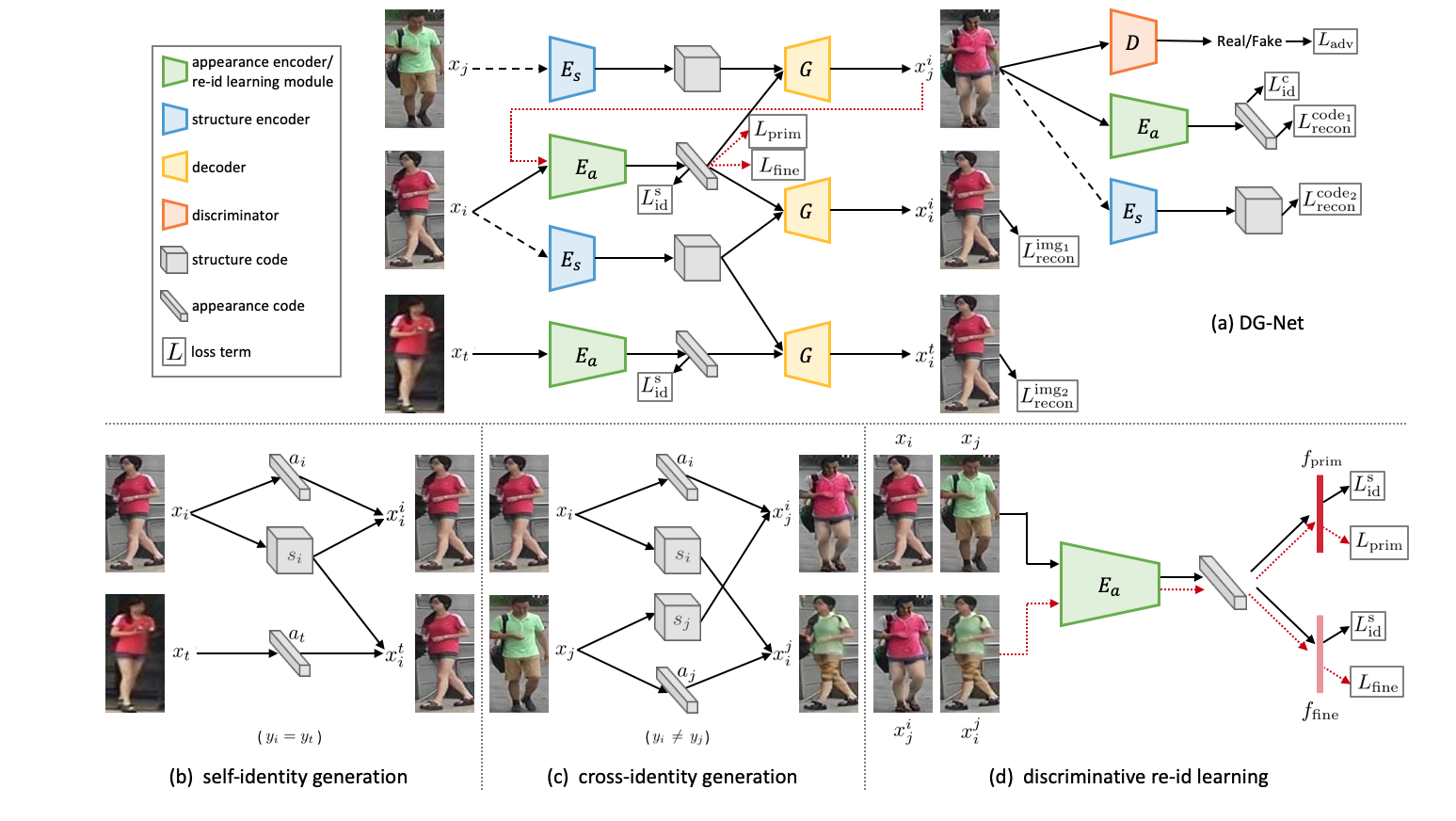}
\end{center}
\vspace{-.23in}
  \caption{ 
  A schematic overview of DG-Net. (a) Our discriminative re-id learning module is embedded in the generative module by sharing appearance encoder $E_a$. A dash black line denotes the input image to structure encoder $E_s$ is converted to gray. The red line indicates the generated images are online fed back to $E_a$. Two objectives are enforced in the generative module: (b) self-identity generation by the same input identity and (c) cross-identity generation by different input identities. (d) To better leverage generated data, the re-id learning involves primary feature learning and fine-grained feature mining.
}
\label{fig:refine}
\end{figure*}

\section{Related Work}

A large family of person re-id research focuses on metric learning loss. Some methods combine identification loss with verification loss \cite{zheng2016discriminatively, wu2018and}, others apply triplet loss with hard sample mining \cite{hermans2017defense, ristani2018features, cheng2016person}. 
Several recent works employ pedestrian attributes to enforce more supervisions and perform multi-task learning \cite{lin2017improving, su2016deep, wang2018transferable}. 
Alternatives harness pedestrian alignment and part matching to leverage on the human structure prior. One of the common practice is to split input images or feature maps horizontally to take advantage of local spatial cues \cite{yi2014deep,li2017person,sun2017beyond}. In a similar manner, pose estimation is incorporated into learning local features \cite{su2017pose, zhao2017spindle, wei2017glad, suh2018part,zheng2018pedestrian}. Apart from pose, human parsing is used in \cite{kalayeh2018human} to enhance spatial matching. In comparison, our DG-Net relies only on simple identification loss for re-id learning and requires no extra auxiliary information such as pose or human parsing for image generation.    

Another active research line is to utilize GANs to augment training data. In \cite{zheng2017unlabeled}, Zheng et al.\ first introduce to use unconditional GAN to generate images from random vectors. Huang et al.\  proceed with this direction with WGAN \cite{arjovsky2017wasserstein} and assign pseudo labels to generated images \cite{huang2018multi}. Li et al.\ propose to share weights between re-id model and discriminator of GAN \cite{li2018adversarial}. In addition, some recent methods make use of pose estimation to conduct pose-conditioned image generation. A two-stage generation pipeline is developed in \cite{ma2017pose} based on pose to refine generated images. Similarly, pose is also used in \cite{ge2018fdgan, liu2018pose,qian2017pose} to generate images of a pedestrian in different poses to make learned features more robust to pose variances. Siarohin et al.\ achieve better pose-conditioned image generation by using a nearest neighbor loss to replace the traditional $\ell_1$ or $\ell_2$ loss \cite{siarohin2018deformable}. All the methods set image generation and re-id learning as two disjointed steps, while our DG-Net end-to-end integrates the two tasks into a unified network.           

Meanwhile, some recent studies also exploit synthetic data for style transfer of pedestrian images to compensate for the disparity between the source and target domains. CycleGAN \cite{CycleGAN2017} is applied in \cite{deng2018image,zhong2019invariance} to transfer pedestrian image style from one dataset to another. StarGAN \cite{choi2018stargan} is used in \cite{zhong2018generalizing} to generate pedestrian images with different camera styles. Bak et al.\  \cite{bak2018domain} employ a game engine to render pedestrians using various illumination conditions. Wei et al.\ \cite{wei2018person} take semantic segmentation to extract foreground mask in assisting style transfer. In contrast to the global style transfer, we aim for manipulating appearance and structure details to facilitate more robust re-id learning. 

\section{Method}

As illustrated in Figure~\ref{fig:refine}, DG-Net tightly couples the generative module for image generation and the discriminative module for re-id learning. We introduce two image mappings: self-identity generation and cross-identity generation 
to synthesize high-quality images that are online fed into re-id learning. Our discriminative module involves primary feature learning and fine-grained feature mining, which are co-designed with the generative module to better leverage the generated data.   

\subsection{Generative Module}
\label{sec:image generator}

\textbf{Formulation.}  
We denote the real images and identity labels as $X=\{x_i\}_{i=1}^N$ and $Y = \{y_i\}_{i=1}^N$, where $N$ is the number of images, $y_i \in [1,K]$ and $K$ indicates the number of classes or identities in the dataset. Given two real images $x_i$ and $x_j$ in the training set, our generative module generates a new pedestrian image by swapping the appearance or structure codes of the two images. As shown in Figure~\ref{fig:refine}, the generative module consists of an appearance encoder $E_a: x_i \rightarrow a_i$, a structure encoder $E_s: x_j \rightarrow s_j$, a decoder $G: (a_i,s_j) \rightarrow x_j^i$, and a discriminator $D$ to distinguish between generated images and real ones. In the case $i = j$, the generator can be viewed as an auto-encoder, so $x^i_i \approx x_i$. Note: for generated images, we use superscript to denote the real image providing appearance code and subscript to indicate the one offering structure code, while real images only have subscript as image index. Compared to the appearance code $a_i$, the structure code $s_j$ maintains more spatial resolution to preserve geometric and positional properties. However, this may result in a trivial solution for $G$ to only use $s_j$ but ignore $a_i$ in image generation since decoders tend to rely on the feature with more spatial information. In practice, we convert input images of $E_s$ into gray-scale to drive $G$ to leverage both $a_i$ and $s_j$. We enforce the two objectives for the generative module: (1) self-identity generation to regularize the generator and (2) cross-identity generation to make generated images controllable and match real data distribution.   

\textbf{Self-identity generation.} 
As illustrated in Figure \ref{fig:refine}(b), given an image $x_i$, the generative module first learns how to reconstruct $x_i$ from itself. This simple self-reconstruction task serves as an important regularization role to the whole generation. We reconstruct the image using the pixel-wise $\ell_1$ loss: 
\begin{equation}
    L^{\mathrm{img_1}}_{\mathrm{recon}} = \mathbb{E}[\left\lVert x_i - G(a_i, s_i) \right\rVert_1].
\end{equation}
Based on the assumption that the appearance codes of the same person in different images are close, we further propose another reconstruction task between any two images of the same identity. In other words, the generator should be able to reconstruct $x_i$ through an image $x_t$ with the same identity $y_i = y_t$: 
\begin{equation}
    L^{\mathrm{img_2}}_{\mathrm{recon}} = \mathbb{E}[\left\lVert x_i - G(a_t, s_i) \right\rVert_1].
\end{equation}
This same-identity but cross-image reconstruction loss encourages the appearance encoder to pull appearance codes of the same identity together so that intra-class feature variations are reduced. In the meantime, to force the appearance codes of different images to stay apart, we use identification loss to distinguish different identities:
\begin{equation}
    L^{\mathrm{s}}_{\mathrm{id}} = \mathbb{E}[-\log(p(y_i|x_i))],
\end{equation}
where $p(y_i|x_i)$ is the predicted probability that $x_i$ belongs to the ground-truth class $y_i$ based on its appearance code.

\textbf{Cross-identity generation.} 
Different from self-identity generation that works with image reconstruction using the same identity, cross-identity generation focuses on image generation with different identities. In this case, there is no pixel-level ground-truth supervision. Instead, we introduce the latent code reconstruction based on appearance and structure codes to control such image generation. As shown in Figure \ref{fig:refine}(c), given two images $x_i$ and $x_j$ of different identities $y_i \neq y_j$, the generated image $x^i_j = G(a_i, s_j)$ is required to retain the information of appearance code $a_i$ from $x_i$ and structure code $s_j$ from $x_j$, respectively. We should then be able to reconstruct the two latent codes after encoding the generated image:
\begin{align}
    L^{\mathrm{code_1}}_{\mathrm{recon}} &= \mathbb{E} [\left\lVert a_i - E_a(G(a_i, s_j)) \right\rVert_1], \\
    L^{\mathrm{code_2}}_{\mathrm{recon}} &= \mathbb{E} [\left\lVert s_j - E_s(G(a_i, s_j)) \right\rVert_1].
\end{align} 
Similar for self-identity generation, we also enforce identification loss on the generated image based on its appearance code to keep the identity consistency: 
\begin{align}
    L^{\mathrm{c}}_{\mathrm{id}} = \mathbb{E}[-\log(p(y_i|x^i_j))],
\end{align}
where $p(y_i|x^i_j)$ is the predicted probability of $x^i_j$ belonging to the ground-truth class $y_i$ of $x_i$, the image that provides appearance code in generating $x^i_j$. Additionally, we employ adversarial loss to match the distribution of generated images to the real data distribution:
\begin{align}
L_{\mathrm{adv}} = \mathbb{E}[\log D(x_i) + \log(1 - D(G(a_i, s_j))].
\end{align}

\textbf{Discussion.} 
By using the proposed generation mechanism, we enable the generative module to learn appearance and structure codes with explicit and complementary meanings and generate high-quality pedestrian images based on the latent codes. This largely eases the generation complexity. In contrast, the previous methods \cite{zheng2017unlabeled, huang2018multi, qian2017pose, ge2018fdgan, liu2018pose} have to learn image generation either from random noise or managing the pose factor only,  
which is hard to manipulate the outputs and inevitably introduces artifacts. Moreover, due to using the latent codes, the variants in our generated images are explainable and constrained in the existing contents of real images, which also ensures the generation realism. In theory, we can generate $O(N\times N)$ different images by sampling various image pairs, resulting in a much larger online generated training sample pool than the ones with $O(2\times N)$ images offline generated in \cite{zheng2017unlabeled, huang2018multi, qian2017pose}. 

\subsection{Discriminative Module}
\label{sec:PersonReID}
Our discriminative module is embedded in the generative module by sharing the appearance encoder as the backbone for re-id learning. In accordance with the images generated by switching either appearance or structure codes, we propose the primary feature learning and fine-grained feature mining to better take advantage of the online generated images. Since the two tasks focus on different aspects of generated images, we branch out two lightweight headers on top of the appearance encoder for the two types of feature learning, as illustrated in Figure~\ref{fig:refine}(d). 

\textbf{Primary feature learning.} 
It is possible to treat the generated images as training samples similar to the existing work~\cite{zheng2017unlabeled, huang2018multi, qian2017pose}. But the inter-class variations in the cross-id composed images motivate us to adopt a teacher-student type supervision with dynamic soft labeling. We use a teacher model to dynamically assign a soft label to $x^i_j$, depending on its compound appearance and structure from $x_i$ and $x_j$. The teacher model is simply a baseline CNN trained with identification loss on the original training set. To train the discriminative module for primary feature learning, we minimize the KL divergence between the probability distribution $p(x^i_j)$ predicted by the discriminative module and the probability distribution $q(x^i_j)$ predicted by the teacher:
\begin{align}
\label{Eq:prim}
    L_{\mathrm{prim}} = \mathbb{E}[ - \sum_{k=1}^K {q(k | x^i_j) \log (\frac{p(k | x^i_j)}{q(k | x^i_j)}) } ],
\end{align}
where $K$ is the number of identities. In comparison with the fixed one-hot label \cite{qian2017pose,zou2018unsupervised} or static smoothing label \cite{zheng2017unlabeled}, this dynamic soft labeling fits better in our case, as each synthetic image is formed by the visual contents from two real images. In the experiments, we show that a simple baseline CNN serving as the teacher model is reliable to provide the dynamic labels and improve the performance.    

\textbf{Fine-grained feature mining.} 
Beyond the direct usage of generated data for learning primary features, an interesting alternative, made possible by our specific generation pipeline, is to simulate the change of clothing for the same person, as shown in each column of Figure~\ref{fig:rainbow}. When training on images organized in this manner, the discriminative module is forced to learn the fine-grained id-related attributes (such as hair, hat, bag, body size, and so on) that are independent to clothing. We view the images generated by one structure code combining with different appearance codes as the same class as the real image providing the structure code. To train the discriminative module for fine-grained feature mining, we enforce identification loss on this particular categorizing:
\begin{align}
\label{Eq:fine}
    L_{\mathrm{fine}} = \mathbb{E}[- \log(p(y_j | x^i_j))].
\end{align}
This loss imposes additional identity supervision to the discriminative module in a multi-tasking way. Moreover, unlike the previous works using manually labeled pedestrian attributes~\cite{lin2017improving, su2016deep, wang2018transferable}, our approach performs automatic fine-grained attribute mining by leveraging on the synthetic images. 
Furthermore, compared to the hard sampling policy applied in \cite{hermans2017defense, ristani2018features}, there is no need to explicitly search for the hard training samples that usually possess fine-grained details, since our discriminative module learns to attention on the subtle identity properties through this fine-grained feature mining.  

\textbf{Discussion.} 
We argue that our high-quality synthetic images, in nature, can be viewed as ``inliers'' (contrary to ``outliers''), as our generated images maintain and recompose the visual contents from real data. Via the above two feature learning tasks, our discriminative module makes specific use of the generated data in line with the way how we manipulate the appearance and structure codes. Instead of using a single supervision as in almost all previous methods \cite{zheng2017unlabeled, huang2018multi, qian2017pose}, we treat the generated images in two different perspectives through the primary feature learning and fine-grained feature mining, where the former focuses on the structure-invariant clothing information and the latter attentions to the appearance-invariant structural cues. 

\subsection{Optimization.} 
We jointly train the appearance and structure encoders, decoder, and discriminator to optimize the total objective, which is a weighted sum of the following losses:
\begin{align}
&L_{\mathrm{total}}(E_a, E_s, G, D) = \lambda_{\mathrm{img}} L^{\mathrm{img}}_{{\mathrm{recon}}} +  L^{\mathrm{code}}_{\mathrm{recon}} \hspace{2pt} + \nonumber\\
&L^{\mathrm{s}}_{\mathrm{id}} + \lambda_{\mathrm{id}} L^{\mathrm{c}}_{\mathrm{id}} + L_{\mathrm{adv}} + \lambda_{\mathrm{prim}}L_{\mathrm{prim}} + \lambda_{\mathrm{fine}}L_{\mathrm{fine}},
\end{align}
where $L^{\mathrm{img}}_{\mathrm{recon}} = L^{\mathrm{img_1}}_{\mathrm{recon}} + L^{\mathrm{img_2}}_{\mathrm{recon}}$ is the image reconstruction loss in self-identity generation, $L^{\mathrm{code}}_{\mathrm{recon}} = L^{\mathrm{code_1}}_{\mathrm{recon}} + L^{\mathrm{code_2}}_{\mathrm{recon}}$ is the latent code reconstruction loss in cross-identity generation, $\lambda_{\mathrm{img}}$, $\lambda_{\mathrm{id}}$, $\lambda_{\mathrm{prim}}$, and $\lambda_{\mathrm{fine}}$ are weights to control the importance of related loss terms. Following the common practice in image-to-image translations~\cite{CycleGAN2017,lee2018diverse, huang2018multimodal}, we use a large weight $\lambda_{\mathrm{img}} = 5$ for the image reconstruction loss. Since the quality of cross-id generated images is not great at the beginning, the identification loss $L^{\mathrm{c}}_{\mathrm{id}}$ may make the training unstable, so we set a small weight $\lambda_{\mathrm{id}} = 0.5$. We fix the two weights during the whole training process in all experiments. We do not involve the discriminative feature learning losses $L_{\mathrm{prim}}$ and $L_{\mathrm{fine}}$ until the generation quality is stable. As an example, we add in the two losses after 30K iterations on Market-1501, then linearly increase $\lambda_{\mathrm{prim}}$ from 0 to 2 in 4K iterations and set $\lambda_{\mathrm{fine}} = 0.2\lambda_{\mathrm{prim}}$. See more details on how to determine the weights in Section~\ref{sec:hyper}. Similar to the alternative updating policy for GANs, in the cross-identity generation as shown in Figure~\ref{fig:refine}(a), we alternatively train $E_a$, $E_s$ and $G$ before the generated image and $E_a$, $E_s$ and $D$ after the generated image. 



\begin{figure*}[t]
\begin{center}
   \includegraphics[width=1.0\linewidth]{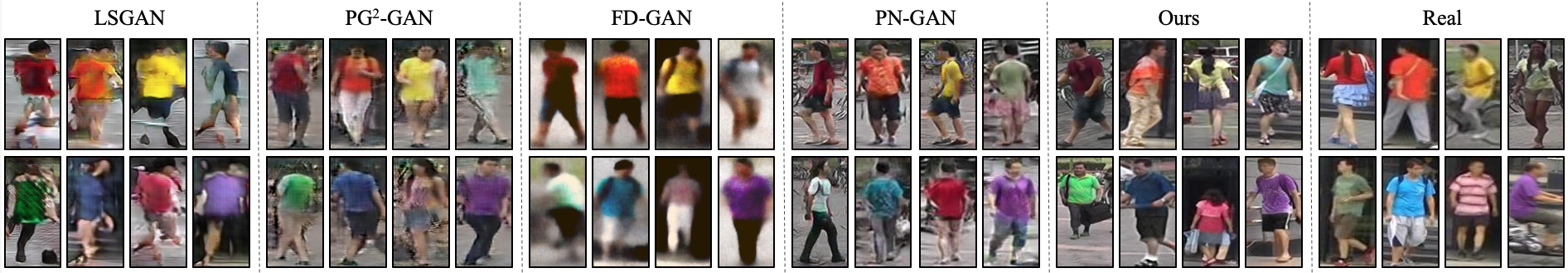}
\end{center}
\vspace{-.2in}
   \caption{Comparison of the generated and real images on Market-1501 across the different methods including LSGAN~\cite{mao2017least}, PG$^2$-GAN~\cite{ma2017pose}, FD-GAN~\cite{ge2018fdgan}, PN-GAN~\cite{qian2017pose}, and our approach. This figure is best viewed when zoom in. Please attention to both foreground and background of the images.} 
\label{fig:gan}
\end{figure*}

\section{Experiments}
We evaluate the proposed approach following standard protocols on three benchmark datasets: Market-1501~\cite{zheng2015scalable}, DukeMTMC-reID~\cite{ristani2016MTMC,zheng2017unlabeled}, and MSMT17~\cite{wei2018person}. We qualitatively and quantitatively compare DG-Net with state-of-the-art methods on both generative and discriminative results. Extensive experiments demonstrate that DG-Net produces more realistic and diverse images, and meanwhile, consistently outperforms the most recent competing algorithms by large margins on re-id accuracy across all benchmarks.       

\begin{figure}[t]
\vspace{-2mm}
\begin{center}
   \includegraphics[width=0.92\linewidth]{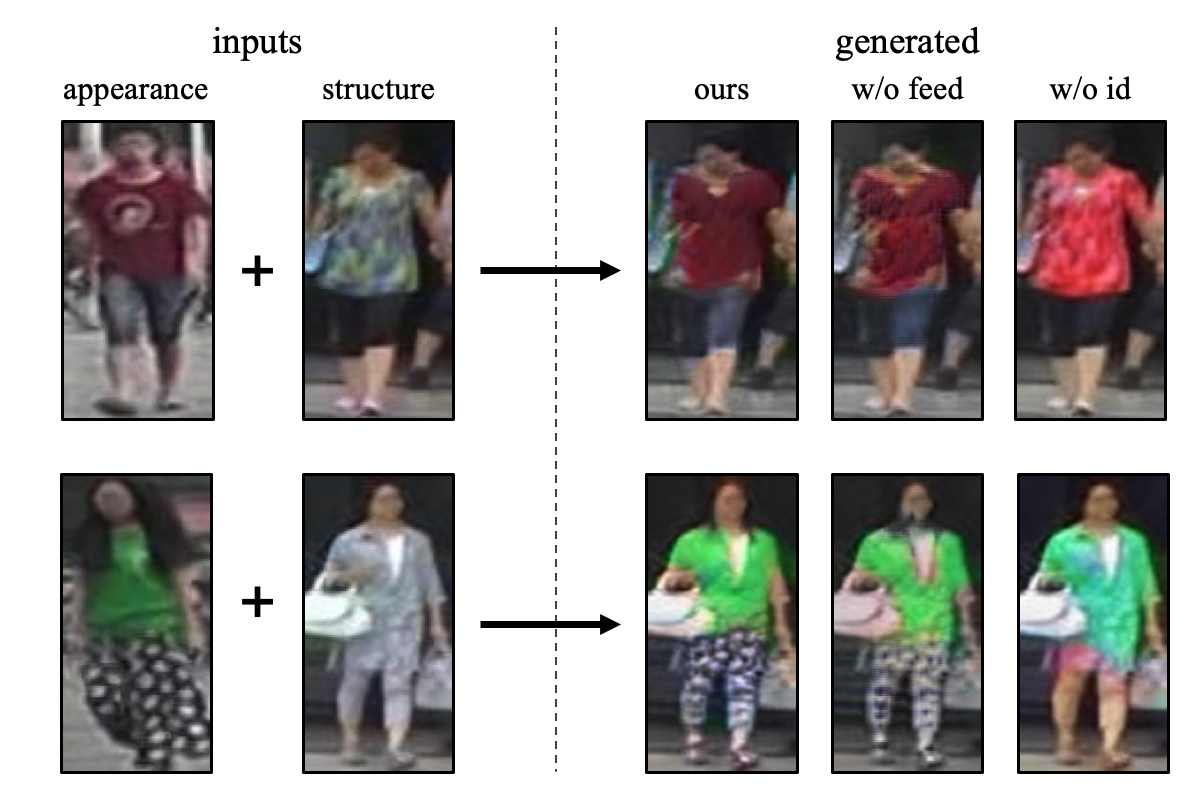}
\end{center}
\vspace{-.2in}
   \caption{Comparison of the generated images by our full model, removing online feeding (w/o feed), and further removing identity supervision (w/o id). 
   }
\label{fig:ablation}
\end{figure}

\subsection{Implementation Details}
\label{sec:Implementation Detail}
Our network is implemented in PyTorch. In the following, we use channel$\times$height$\times$width to indicate the size of feature maps. (\textbf{\romannumeral 1}) $E_a$ is based on ResNet50~\cite{he2016deep} pre-trained on ImageNet~\cite{imagenet}, and we remove its global average pooling layer and fully-connected layer then append an adaptive max pooling layer to output the appearance code $a$ in $2048\times4\times1$. It is mapped to primary feature $f_{\mathrm{prim}}$ and fine-grained feature $f_{\mathrm{fine}}$, both are 512-dim vectors, through two fully-connected layers. 
(\textbf{\romannumeral 2}) $E_s$ is a shallow network that outputs the structure code $s$ in $128\times64\times32$. It consists of four convolutional layers followed by four residual blocks~\cite{he2016deep}. 
(\textbf{\romannumeral 3}) $G$ processes $s$ by four residual blocks and four convolutional layers. As in~\cite{huang2018multimodal} every residual block contains two adaptive instance normalization layers~\cite{huang2017arbitrary}, which integrate in $a$ as scale and bias parameters.
(\textbf{\romannumeral 4}) $D$ follows the popular multi-scale PatchGAN~\cite{isola2017image}. We employ discriminators on the three different input image scales: $64\times32$, $128\times64$, and $256\times128$. We also apply the gradient punishment~\cite{Mescheder2018ICML} when updating $D$ to stabilize training. 
(\textbf{\romannumeral 5}) For training, all input images are resized to $256\times128$. Similar to the previous deep re-id models~\cite{zheng2016survey}, SGD is used to train $E_a$ with learning rate $0.002$ and momentum $0.9$. We apply Adam~\cite{kingma2014adam} to optimize $E_s$, $G$ and $D$, and set learning rate to $0.0001$, and $(\beta_1, \beta_2) = (0, 0.999)$. 
(\textbf{\romannumeral 6}) At test time, our re-id model only involves $E_a$ (along with two lightweight headers), which is of a comparable network size to most methods using ResNet50 as the backbone. We concatenate $f_{\mathrm{prim}}$ and 
$f_{\mathrm{fine}}$ into a 1024-dim vector as the final pedestrian representation. 
More architecture details can be found in the appendix.


\begin{figure}[t]
\vspace{-2mm}
\begin{center}
   \includegraphics[width=1\linewidth]{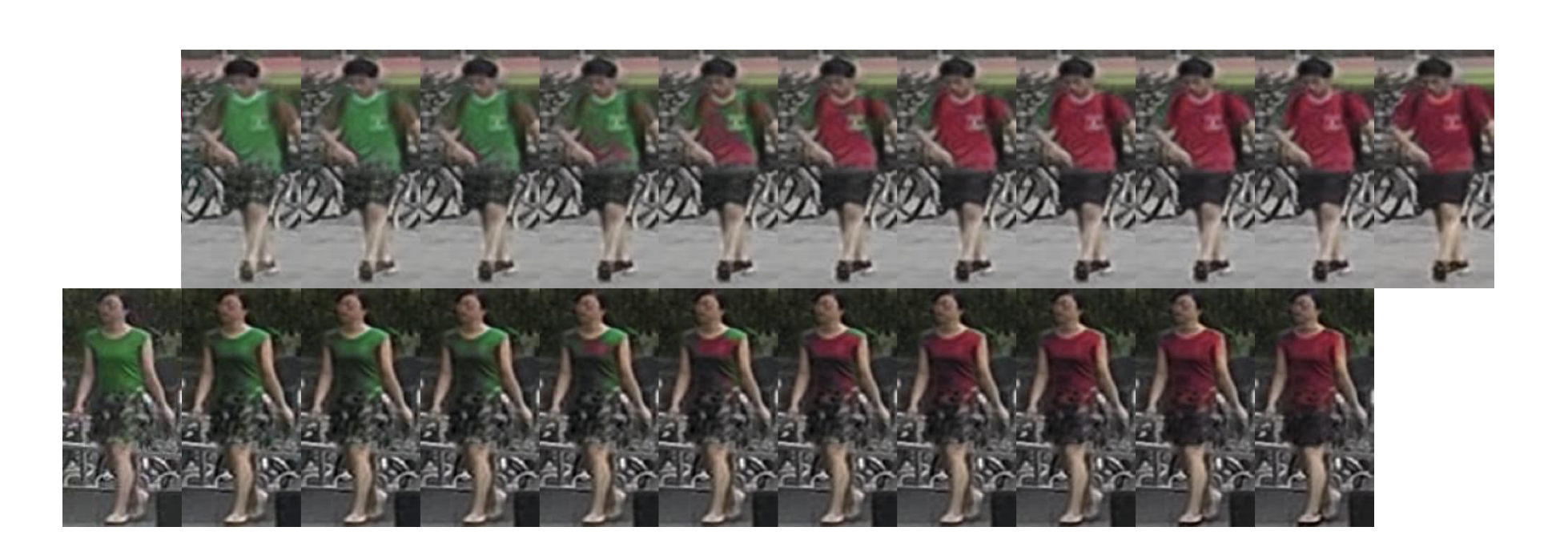}
\end{center}
\vspace{-.2in}
   \caption{Example of image generation by linear interpolation between two appearance codes. 
   }
\label{fig:smooth}
\end{figure}

\subsection{Generative Evaluations}

\textbf{Qualitative evaluations.}
We first qualitatively compare DG-Net with its two variants that ablate online feeding and identity supervision. As shown in Figure~\ref{fig:ablation}, without online feeding generated images to appearance encoder, the model suffers from blurry edges and undesired textures.
If further removing identity supervision, the image quality is unsatisfying as the model fails to produce the accurate clothing color or style. This clearly shows that our joint discriminative learning is beneficial to the image generation. 

\begin{figure*}[tbh]
\begin{center}
   \includegraphics[width=0.95\linewidth]{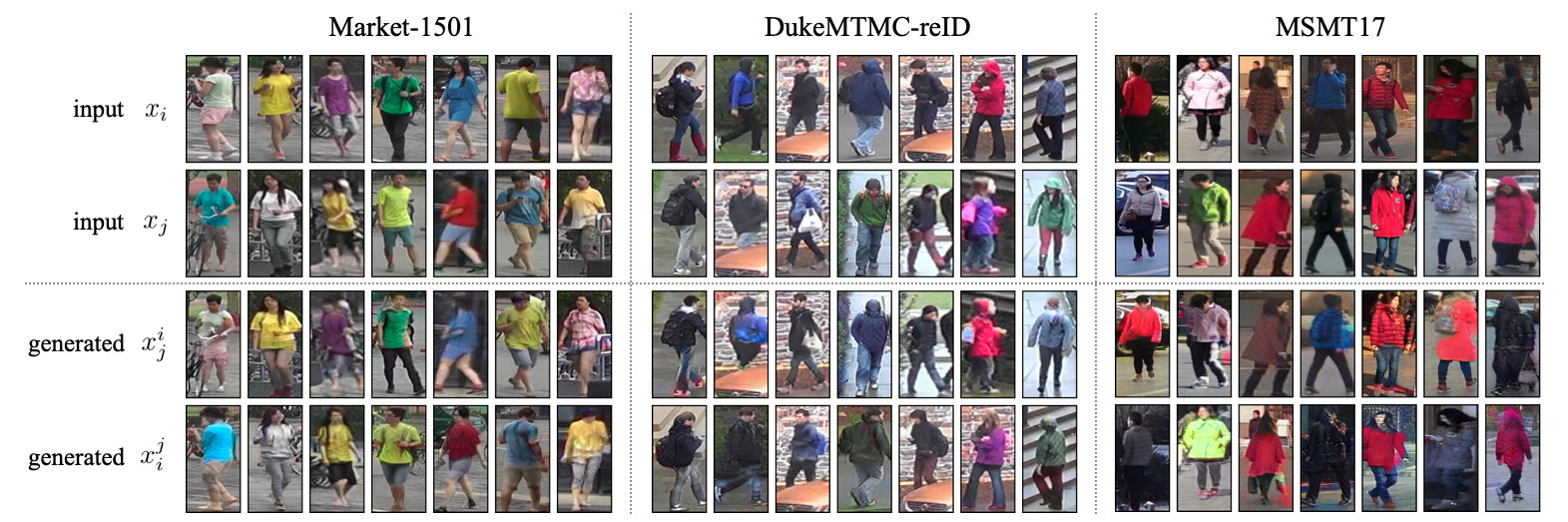}
\end{center}
\vspace{-.2in}
   \caption{
   Examples of our generated images by swapping appearance or structure codes on the three datasets. All images are sampled from the test sets.}
\label{fig:swap}
\end{figure*}

Next we compare our full model with other generative approaches, including one unconditional GAN (LSGAN \cite{mao2017least}) and three open-source conditional GANs (PG$^2$-GAN \cite{ma2017pose}, PN-GAN \cite{qian2017pose} and FD-GAN \cite{ge2018fdgan}). As compared in Figure~\ref{fig:gan}, the images generated by LSGAN have severe artifacts and duplicated patterns. FD-GAN are prone to generate very blurry images, which largely deteriorate the realism. PG$^2$-GAN and PN-GAN, both conditioned on pose, generate relatively good visual results, but still contain visible blurs and artifacts especially in background. In comparison, our generated images are more realistic and close to the real in both foreground and background. 

To better understand the learned appearance space, which is the foundation for our pedestrian representations, we perform a linear interpolation between two appearance codes and generate the corresponding images as shown in Figure~\ref{fig:smooth}. These interpolation results verify the continuity in the appearance space, and show that our model is able to generalize in the space instead of simply memorizing trivial visual information. As a complementary study, we also generate images by linearly interpolating between two structure codes while keeping the appearance code intact. See more discussions regarding this study in the appendix. We then demonstrate our generation results on the three benchmarks in Figure~\ref{fig:swap}, where DG-Net is found to be able to consistently generate realistic and diverse images across the different datasets.

\begin{table}[tbp]
\vspace{-2.5mm}
\small
\caption{Comparison of FID (lower is better) and SSIM (higher is better) to evaluate realism and diversity of the real and generated images on Market-1501.} 
{
\label{table:visual}
\setlength{\tabcolsep}{15pt}
\begin{tabular}{l|c|c}
\shline
\multirow{2}{*}{Methods} & \multicolumn{1}{c|}{Realism}  & \multicolumn{1}{c}{Diversity}  \\
 & (FID) & (SSIM) \\
\hline
Real & 7.22 & 0.350 \\ 
\hline
LSGAN \cite{mao2017least} & 136.26 & - \\
PG$^2$-GAN \cite{ma2017pose} & 151.16 & - \\
PN-GAN \cite{qian2017pose} & 54.23 & 0.335\\
FD-GAN \cite{ge2018fdgan} & 257.00 &  0.247\\ 
\hline
Ours & \textbf{18.24} & \textbf{0.360} \\
\shline
\end{tabular}}
\end{table}

\begin{figure}[t]
\vspace{-2.5mm}
\begin{center}
   \includegraphics[width=1\linewidth]{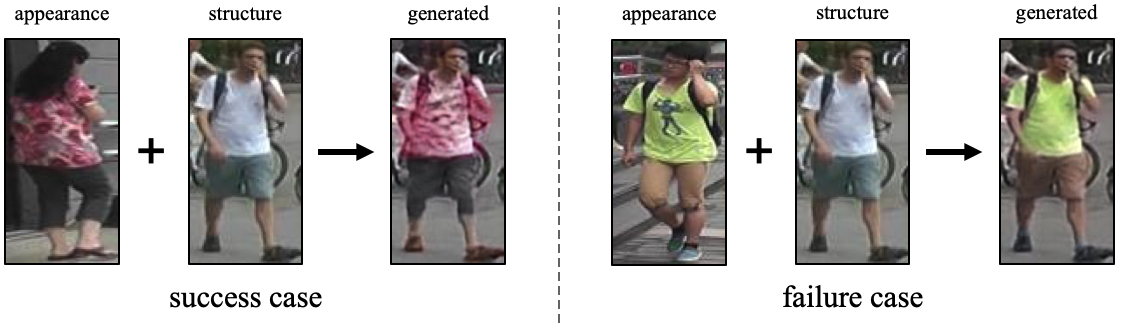}
\end{center}
\vspace{-.2in}
   \caption{Comparison of success and failure cases in our image generation. In the failure case, the logo on t-shirt of the original image is missed in the synthetic image. 
   }
\label{fig:fail}
\end{figure}

\textbf{Quantitative evaluations.}
Our qualitative observations above are confirmed by the quantitative evaluations. We use two metrics: Fr\'echet Inception Distance (FID)\cite{heusel2017gans} and Structural SIMilarity (SSIM) \cite{wang2004image} to measure realism and diversity of generated images, respectively. FID measures how close the distribution of generated images is to the real. It is sensitive to visual artifacts and thus indicates the realism of generated images. For the identity conditioned generation, we apply SSIM to compute intra-class similarity, which can be used to reflect the generation diversity. As shown in Table~\ref{table:visual}, our approach significantly outperforms other methods on both realism and diversity, suggesting the high quality of our generated images. Remarkably, we obtain a higher SSIM than the original training set thanks to the various poses, carryings, backgrounds, etc. introduced by switching structure codes.

\textbf{Limitation.} 
We notice that due to data bias in the original training set, our generative module tends to learn the regular textures (e.g., stripes and dots) but ignores some rare patterns (e.g., logos on shirts), as shown in Figure~\ref{fig:fail}.  

\subsection{Discriminative Evaluations}
\textbf{Ablation studies.} 
We first study the contributions of primary feature and fine-grained feature in Table~\ref{table:ablation_s}. We train ResNet50 with identification loss on each original training set as the baseline. It also serves as the teacher model in primary feature learning to perform dynamic soft labeling on the generated images. Our primary feature is found to largely improve over the baseline. Notably, the fine-grained feature without using important appearance information but only considering subtle id-related cues already achieves impressive accuracy. By combining the two features, we can further improve the performance, which substantially outperforms the baseline by $6.1\%$ for Rank@1 and $12.4\%$ for mAP on average of the three datasets. We then evaluate the two features independently learned after our synthetic images are offline generated. This results in an $84.4\%$ mAP on Market-1501, inferior to the $86.0\%$ mAP of the end-to-end training, suggesting that our joint generative training is beneficial to the re-id learning. 

\begin{table}[t]
\small
\vspace{-.1in}
\caption{ 
Comparison of baseline, primary feature, fine-grained feature, and their combination on the three datasets.}
{
\label{table:ablation_s}
\setlength{\tabcolsep}{1pt}
\begin{tabular}{l|cc|cc|cc}
\shline
\multirow{2}{*}{Methods} & \multicolumn{2}{c|}{Market-1501} & \multicolumn{2}{c|}{DukeMTMC-reID} &\multicolumn{2}{c}{MSMT17}\\
& Rank@1 & mAP & Rank@1  & mAP & Rank@1  & mAP\\
\hline
Baseline & 89.6 & 74.5 & 82.0  & 65.3 & 68.8 & 36.2 \\
$f_{\mathrm{prim}}$ & 94.0 & 84.4 & 85.6 & 72.7 &  76.0 & 49.7 \\
$f_{\mathrm{fine}}$  & 91.6 & 75.3 & 78.7 & 61.2 & 71.5 & 43.5 \\
\hline
$f_{\mathrm{prim}}, f_{\mathrm{fine}}$ & \textbf{94.8} & \textbf{86.0} & \textbf{86.6} & \textbf{74.8} & \textbf{77.2} & \textbf{52.3} \\
\shline
\end{tabular}}
\end{table}

\begin{figure}[t]
\begin{center}
   \includegraphics[width=1\linewidth]{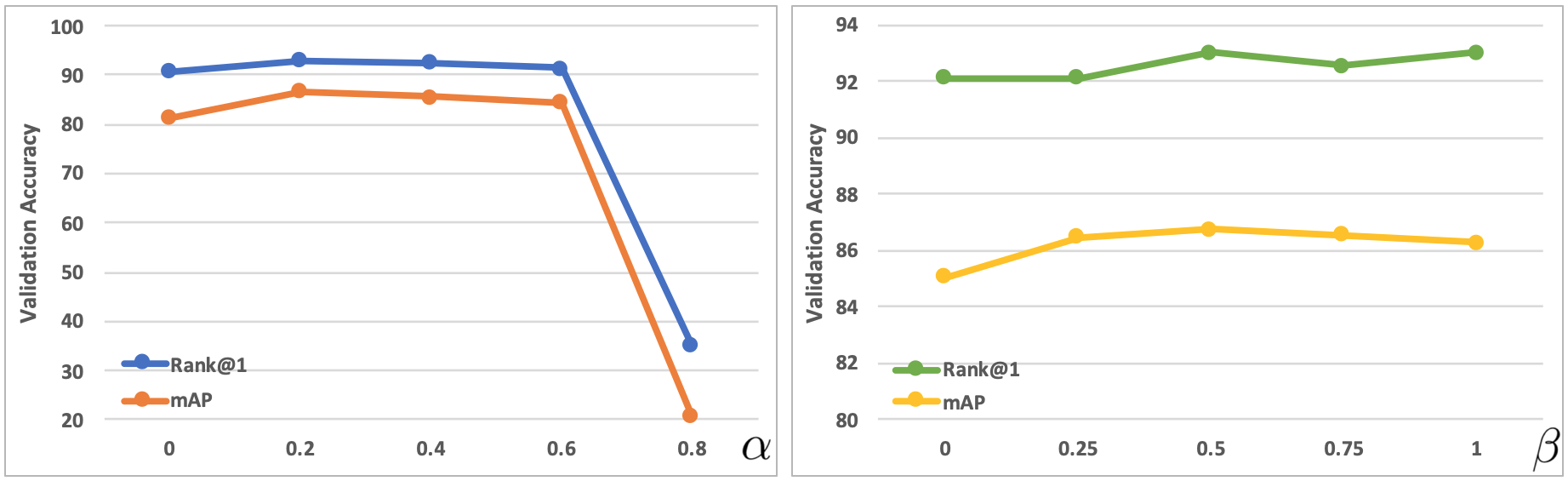}
\end{center}
\vspace{-.2in}
   \caption{Analysis of the re-id learning related hyper-parameters $\alpha$ and $\beta$ to balance primary and fine-grained features in training (left) and testing (right).  
   }
\label{fig:hyper}
\end{figure}

\textbf{Influence of hyper-parameters.}
\label{sec:hyper}
Here we show how to set the re-id learning related weights: one is $\alpha$, the ratio between $\lambda_{\mathrm{fine}}$ and $\lambda_{\mathrm{prim}}$ to control the importance of $L_{\mathrm{fine}}$ and $L_{\mathrm{prim}}$ in training; the other is $\beta$ to weight $f_\mathrm{fine}$ when combined with $f_\mathrm{prim}$ as the final pedestrian representation in testing. We search the two hyper-parameters on a validation set split out from the original training set of Market-1501 (first 651 classes for training and rest 100 classes for validation). Based on the valiation results in Figure~\ref{fig:hyper}, we choose $\alpha = 0.2$ and $\beta = 0.5$ in all experiments.

\textbf{Comparison with state-of-the-art methods.}
Finally we report the performance of our approach with other state-of-the-art results in Tables~\ref{table:mr} and~\ref{table:msmt}. Note that we do not apply any post processing such as re-ranking~\cite{yu2017divide} or multi-query fusion~\cite{zheng2015scalable}. On each dataset, our approach attains the best performance. Comparing with the methods using separately generated images, DG-Net achieves clear gains of $8.3\%$ and $10.3\%$ for mAP on Market-1501 and DukeMTMC-reID, indicating the advantage of the proposed joint learning. Moreover, our framework is more training efficient: we use only one training phase for joint image generation and re-id learning, while others require two training phases to sequentially train generative models and re-id models. DG-Net also outperforms other non-generative methods by large margins on the two datasets. As for the recent released large-scale dataset MSMT17, DG-Net performs significantly better than the second best method by $9.0\%$ for Rank@1 and $11.9\%$ for mAP.           

\begin{table}[tbp]
\small
\vspace{-.1in}
\caption{Comparison with the state-of-the-art methods on the Market-1501 and DukeMTMC-reID datasets. Group 1: the methods not using generated data. Group 2: the methods using separately generated images.
}
{
\label{table:mr}
\setlength{\tabcolsep}{6pt}
\begin{tabular}{l|cc|cc}
\shline
\multirow{2}{*}{Methods} & \multicolumn{2}{c|}{Market-1501} & \multicolumn{2}{c}{DukeMTMC-reID}\\
& Rank@1 & mAP & Rank@1  & mAP \\
\hline
Verif-Identif \cite{zheng2016discriminatively} & 79.5 & 59.9 & 68.9 & 49.3 \\
DCF \cite{Li2017Learning}  & 80.3 & 57.5 & - & -\\
SSM \cite{bai2017scalable}  & 82.2 & 68.8 & - & -\\
SVDNet \cite{sun2017svdnet} & 82.3 & 62.1 & 76.7 & 56.8\\
PAN \cite{zheng2018pedestrian} & 82.8 & 63.4 & 71.6 & 51.5 \\
GLAD \cite{wei2017glad} & 89.9 & 73.9 & - & - \\
HA-CNN \cite{li2018harmonious} & 91.2 & 75.7 & 80.5 & 63.8 \\
MLFN \cite{chang2018multi} & 90.0 & 74.3 & 81.0 & 62.8 \\
Part-aligned \cite{suh2018part} & 91.7 & 79.6 & 84.4 & 69.3 \\
PCB \cite{sun2017beyond} & 93.8 & 81.6 & 83.3 & 69.2\\
Mancs \cite{wang2018mancs} & 93.1 & 82.3 & 84.9 & 71.8 \\
\hline
DeformGAN \cite{siarohin2018deformable} & 80.6 & 61.3 & - & -  \\
LSRO \cite{zheng2017unlabeled} & 84.0 & 66.1 & 67.7 & 47.1  \\
Multi-pseudo \cite{huang2018multi} & 85.8 & 67.5 & 76.8 & 58.6  \\
PT \cite{liu2018pose} & 87.7 & 68.9 & 78.5 & 56.9  \\ 
PN-GAN \cite{qian2017pose} & 89.4 & 72.6 & 73.6 & 53.2 \\
FD-GAN \cite{ge2018fdgan} & 90.5 & 77.7 & 80.0 & 64.5 \\
\hline
Ours &  \textbf{94.8} & \textbf{86.0} & \textbf{86.6} & \textbf{74.8} \\
\shline
\end{tabular}}
\end{table}

\begin{table}[tbp]
\small
{
\setlength{\tabcolsep}{5pt}
\begin{tabular}{l|c|c|c|c}
\shline
Methods & Rank@1 & Rank@5 & Rank@10 & mAP \\
\hline
Deep \cite{szegedy2015going}  & 47.6 & 65.0 & 71.8 & 23.0 \\
PDC \cite{su2017pose} & 58.0 & 73.6 & 79.4 & 29.7 \\
Verif-Identif \cite{zheng2016discriminatively} & 60.5 & 76.2 & 81.6 & 31.6 \\ 
GLAD \cite{wei2017glad} & 61.4 & 76.8 & 81.6 & 34.0 \\
PCB \cite{sun2017beyond} & 68.2 & 81.2 & 85.5 & 40.4\\
\hline
Ours & \textbf{77.2} & \textbf{87.4} & \textbf{90.5} & \textbf{52.3} \\
\shline
\end{tabular}}
\caption{Comparison with the state-of-the-art methods on the MSMT17 dataset.}
\label{table:msmt}
\end{table}


\section{Conclusion}
In this paper, we have proposed a joint learning framework that end-to-end couples re-id learning and image generation in a unified network. There exists an online interactive loop between the discriminative and generative modules to mutually benefit the two tasks. Our two modules are co-designed to let the re-id learning better leverage the generated data, rather than simply training on them. Experiments on three benchmarks demonstrate that our approach consistently brings substantial improvements to both image generation quality and re-id accuracy.     


{\footnotesize
\bibliographystyle{ieee_fullname}
\bibliography{egbib}
}

\clearpage

\appendix
\section*{Appendix}
\setcounter{section}{0}
\renewcommand\thesection{\Alph{section}}

\newcommand{\resblock}[2]{\multirow{3}{*}{\(\left[\begin{array}{c}\text{3$\times$3, #1}\\[-.1em] \text{3$\times$3, #1} \end{array}\right]\)$\times$#2}
}

\newcommand{\convblock}[2]{\multirow{1}{*}{\([\begin{array}{c}\text{#2$\times$#2, #1} \end{array}]\)}
}

\newcommand{\asppblock}[2]{\multirow{3}{*}{\(\left[\begin{array}{c}\text{1$\times$1, #1}\\[-.1em] \text{3$\times$3, #1} \end{array}\right]\)$\times$#2}
}

In this appendix, Section~\ref{sec:architecture} summarizes the architecture details of DG-Net. Section~\ref{sec:more} presents more re-id evaluations. Section~\ref{sec:space} provides more rationales behind the appearance and structure spaces as well as the primary and fine-grained feature learning on appearance code.    
Section~\ref{sec:interpolate} demonstrates the example of image generation by interpolating between structure codes.

\section{Network Architectures}
\label{sec:architecture}
Our proposed DG-Net consists of the appearance encoder $E_a$, structure encoder $E_s$, decoder $G$, and discriminator $D$. As described in the paper that $E_a$ is modified from ResNet50, we now introduce the architecture details of $E_s$, $G$, and $D$. Following the common practice in GANs, 
we mainly adopt convolutional layers and residual blocks \cite{he2016deep} to construct them.

Table \ref{tab:es} shows the architecture of $E_s$. After each convolutional layer, we apply the instance normalization layer \cite{ulyanov2016instance} and LReLU (negative slope set to 0.2). We also add the optional atrous spatial pyramid pooling (ASPP) \cite{chen2017rethinking}, which contains dilated convolutions and can be used to exploit multi-scale features.  
Table \ref{tab:g} demonstrates the architecture of decoder $G$, 
which involves several residual blocks followed by upsampling and convolutional layers. Similar to \cite{huang2018multimodal}, we insert the adaptive instance normalization (AdaIN) layer in every residual block to integrate the appearance code from $E_a$ as the dynamically generated weight and bias parameters of AdaIN. We employ the multi-scale PatchGAN \cite{CycleGAN2017} as the descriminator $D$. Given an input image of $256\times128$, we resize the image to the three different scales: $256\times128$,  $128\times64$, $64\times32$ before feeding them into the discriminator. LReLU (negative slope set to 0.2) is applied after each convolutional layer. We present the architecture of $D$ in Table \ref{tab:d}. 


\section{More Discriminative Evaluations}
\label{sec:more}
In order to have a more thorough evaluation of our approach, we further evaluate the performance of DG-Net on a relatively small dataset. So we generalize our approach to CUHK03-NP \cite{zhong2017re}, which contains much fewer images (9.6 training images per person on average) compared to Market-1501 \cite{zheng2015scalable}, DukeMTMC-reID \cite{ristani2016MTMC} and MSMT17 \cite{wei2018person}. As compared in Table \ref{table:cuhk03-np}, DG-Net achieves 65.6\% Rank@1 and 61.1\% mAP.

\begin{table}[t]
\caption{Architecture of the structure encoder $E_s$.} 
\label{tab:es}
{
\setlength{\tabcolsep}{10pt}
\begin{tabular}{l|c|c}
\shline
Layer & Parameters & Output Size \\
\hline 
Input & - & 1 $\times$ 256 $\times$ 128 \\
\hline
Conv1 & \convblock{16}{3} & 16 $\times$ 128 $\times$ 64 \\
Conv2 & \convblock{32}{3} & 32 $\times$ 128 $\times$ 64 \\
Conv3 & \convblock{32}{3}  & 32 $\times$ 128 $\times$ 64 \\
Conv4 & \convblock{64}{3}  & 64 $\times$ 64 $\times$ 32 \\
\hline
\multirow{3}{*}{ResBlocks} & \resblock{64}{4} & \multirow{3}{*}{ 64 $\times$ 64 $\times$ 32}\\
  &  &  \\
  &  &  \\
\hline
\multirow{4}{*}{ASPP} &  \convblock{32}{1}  & \multirow{4}{*}{128 $\times$ 64 $\times$ 32} \\
  & \asppblock{32}{3} &  \\
  &  &  \\
  &  &  \\
\hline
Conv5 &  \convblock{128}{1} & 128 $\times$ 64 $\times$ 32 \\
\shline
\end{tabular}}
\end{table}

\begin{table}[h]
\caption{Architecture of the decoder $G$.} 
\label{tab:g}
{
\setlength{\tabcolsep}{8pt}
\begin{tabular}{l|c|c}
\shline
Layer & Parameters & Output Size \\
\hline 
Input & - & 128 $\times$ 64 $\times$ 32 \\
\hline
 \multirow{3}{*}{ResBlocks} & \resblock{128}{4} & \multirow{3}{*}{ 128 $\times$ 64 $\times$ 32}\\
  &  &  \\
  &  &  \\
\hline
Upsample & - & 128 $\times$ 128 $\times$ 64 \\
Conv1 &  \convblock{64}{5}  & 64 $\times$ 128 $\times$ 64 \\
\hline
Upsample & - & 64 $\times$ 256 $\times$ 128 \\
Conv2 &  \convblock{32}{5}  & 32 $\times$ 256 $\times$ 128 \\
\hline
Conv3 & \convblock{32}{3} & 32 $\times$ 256 $\times$ 128 \\
Conv4 & \convblock{32}{3} & 32 $\times$ 256 $\times$ 128 \\
Conv5 & \convblock{3}{1} & 3 $\times$ 256 $\times$ 128 \\
\shline
\end{tabular}}
\end{table}

\begin{table}[h]
\caption{Architecture of the discriminator $D$.}
\label{tab:d}
{
\setlength{\tabcolsep}{8pt}
\begin{tabular}{l|c|c}
\shline
Layer & Parameters & Output Size \\
\hline 
Input & - & 3 $\times$ 256 $\times$ 128 \\
\hline
Conv1 & \convblock{32}{1} & 32 $\times$ 256 $\times$ 128 \\
Conv2 & \convblock{32}{3} & 32 $\times$ 256 $\times$ 128 \\
Conv3 & \convblock{32}{3}  & 32 $\times$ 128 $\times$ 64 \\
Conv4 & \convblock{32}{3}  & 32 $\times$ 128 $\times$ 64 \\
Conv5 & \convblock{64}{3}  & 64 $\times$ 64 $\times$ 32 \\
\hline
\multirow{3}{*}{ResBlocks} & \resblock{64}{4} & \multirow{3}{*}{ 64 $\times$ 64 $\times$ 32}\\
  &  &  \\
  &  &  \\
\hline
Conv6 &  \convblock{1}{1} & 1 $\times$ 64 $\times$ 32 \\
\shline
\end{tabular}}
\end{table}

\begin{table*}
\begin{center}
{
\setlength{\tabcolsep}{5pt}
\begin{tabular}{l|c|c|c|c|c|c}
\shline
 & Market$\rightarrow$Duke & Duke$\rightarrow$Market& Market$\rightarrow$MSMT & MSMT$\rightarrow$Market & Duke$\rightarrow$MSMT & MSMT$\rightarrow$Duke \\
\hline
 Rank@1  & 42.62\%   & 56.12\%         & 17.11\%         & 61.76\%         & 20.59\% & 61.89\%         \\
 Rank@5  & 58.57\%   & 72.18\%         & 26.66\%         & 77.67\%         & 31.67\%         & 75.81\%         \\
 Rank@10 & 64.63\%   & 78.12\%         & 31.62\%         & 83.25\%         & 37.04\%         & 80.34\%         \\
 mAP     & 24.25\%   & 26.83\%         & 5.41\%          & 33.62\%         & 6.35\%          & 40.69\%         \\
\shline
\end{tabular}}
\end{center}
\vspace{-.1in}
\caption{Direct transfer learning results.}
\label{table:transfer}
\end{table*}

\section{Appearance and Structure Codes}
\label{sec:space}
Since we cannot quantitatively justify the attributes of appearance/structure codes, Table 1 in the paper is used to qualitatively give an intuition. Our design of $E_s$ (a shallow network) makes the structure space primarily preserve the structural information, such as position and geometry of humans and objects. Thus, the structure code is mainly used to hold the low-level positional and geometric information, such as pose and background that are non-id-related, to facilitate image synthesis. 
On the other hand, certain structure cues, such as bag/hair/body outline, are clearly id-related and are better to be captured by the discriminative module. However, softmax loss is generally too ``lazy'' to be able to capture useful structure information besides appearance features, therefore, the goal of fine-grained feature mining upon the appearance code promotes mining the id-related semantics out of structure cues, also guarantees the complementary nature between primary and fine-grained features.

\section{Interpolate between Structure Codes}
\label{sec:interpolate}
\begin{figure}
\begin{center}
   \includegraphics[width=\linewidth]{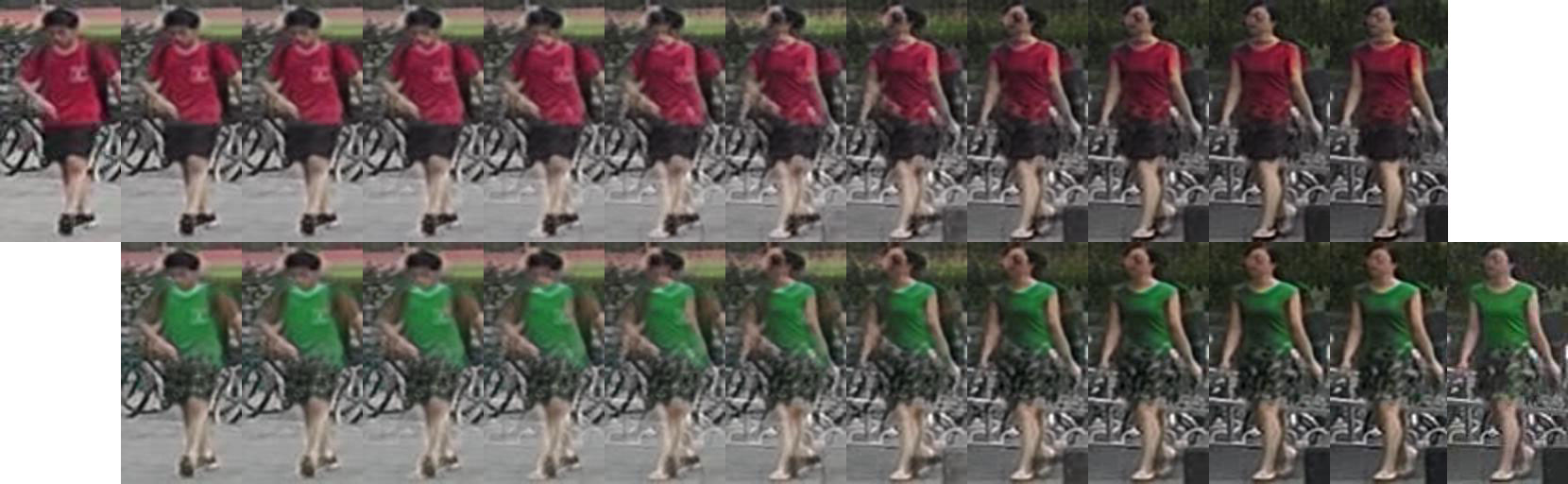}
\end{center}
\vspace{-.1in}
   \caption{Example of image generation by linear interpolation of two structure codes. We fix the appearance code in each row. This figure is best viewed when zoom in and compare with Figure \ref{fig:smooth}. }
\label{fig:structure-interpolation}
\end{figure}

Figure 5 in the paper shows the examples of synthesized images by linear interpolation between two appearance codes. This qualitatively validates the continuity in the appearance space. As a complementary study, here we generate the images by linearly interpolating between two structure codes while keeping the appearance codes intact in Figure~\ref{fig:structure-interpolation}. This demonstrates the exact opposite setting to Figure 5. As expected, most images (both foreground and background) look not realistic. Our hypothesis is that the structure codes are extracted by a shallow network and contain the positional and geometric information of inputs. So the interpolation between the low-level features is not able to preserve semantic smoothness or consistency. 

\section{Direct transfer learning}
To verify the generalizability of DG-Net, we train the model on dataset A and directly test the model on dataset B (with no adaptation). We denote the direct transfer learning protocol as A$\rightarrow$B. The results are shown in Table~\ref{table:transfer}.

\vspace{3mm}

\noindent\textbf{Acknowledgement.} Yi Yang acknowledges support from Data to Decision Cooperative Research Centre. 

\begin{table}
\begin{center}
{
\setlength{\tabcolsep}{15pt}
\begin{tabular}{l|c|c}
\shline
Methods & Rank@1 & mAP \\
\hline
HA-CNN \cite{li2018harmonious} & 41.7\% & 38.6\% \\
PT \cite{liu2018pose} & 41.6\% & 38.7\% \\
MLFN \cite{chang2018multi} & 52.8\% & 47.8\% \\
PCB \cite{sun2017beyond} & 61.3\% & 54.2\% \\
PCB + RPP \cite{sun2017beyond} & 63.7\% & 57.5\% \\
\hline
Ours & \textbf{65.6\%} & \textbf{61.1\%} \\
\shline
\end{tabular}}
\end{center}
\vspace{-.1in}
\caption{Comparison with the state-of-the-art results on the CUHK03-NP dataset.}
\label{table:cuhk03-np}
\end{table}

\end{document}